\documentclass[a4,11pt]{article}
\usepackage{amsthm}
\usepackage{amsmath}
\usepackage{amssymb}
\usepackage{natbib}
\usepackage{booktabs} 
\usepackage{fullpage}
\usepackage{xcolor}
\usepackage{enumitem}

\newif\ifpdf
\ifx\pdfoutput\undefined
  \pdffalse                     
  \usepackage[dvips]{graphicx}
\else
  \pdfoutput=1                  
  \pdftrue
  \usepackage[pdftex]{graphicx}
\fi
  
\usepackage[scaled]{helvet}
\usepackage[T1]{fontenc}

\newcommand{\bfx}{\mathbf{x}}

\def\cut#1{}

\newcommand{\rtrain}{r_{\mathrm{train}}}

\title{
The Effect of Class Imbalance on Precision-Recall Curves}

\author{Christopher K I Williams \\ School of Informatics, University
  of Edinburgh, UK}

\date{\today}

\begin{document}
\maketitle

\begin{abstract}
  In this note I study how the precision of a binary classifier depends on the
  ratio $r$ of positive to negative cases in the test set, as well as
  the classifier's true and false positive rates. This relationship
  allows prediction of how the precision-recall curve will change with
  $r$, which seems not to be well known. It also allows prediction of
  how $F_{\beta}$ and the Precision Gain and Recall Gain measures of
  \cite{flach-kull-15} vary with $r$.
 \end{abstract}

Consider a binary classifier, where the predictions change as the
threshold for deciding between the two classes is varied.  The
Receiver Operating Characteristic (or ROC) curve and the
Precision-Recall (PR) curve are two ways of summarizing the
performance of classifier in this situation. The ROC curve is
invariant to the ratio $r$ of positive to negative cases in the test
set in the population limit, but the PR curve \emph{is} affected by
$r$. Below I show explicitly how the PR curve and
derived quantities like the $F_{\beta}$ measure (due to
\citealt*{van-rijsbergen-79}) are affected by $r$. As these are
frequently used to assess the performance of classifiers, it is
important that the effect of $r$ is well understood, and 
adjusted for (if necessary).

The standard notation (see e.g.,
\citealt*[sec.\ 5.8]{witten-frank-hall-pal-17}) for binary
classification is summarized below:

\vspace*{3mm}
\begin{center}
\begin{tabular}{l|cc|c}
     & \multicolumn{2}{c|}{Predicted}  & Sum \\ \hline
    Actual & positive  & negative & \\ \hline
    positive & $\mathrm{TP}$        & $\mathrm{FN}$ &    $\mathrm{P}$ \\
    negative & $\mathrm{FP}$        & $\mathrm{TN}$ &    $\mathrm{N}$ \\ \hline
\end{tabular}
\end{center}
\vspace*{2mm}

There are $\mathrm{P}$ positive and $\mathrm{N}$ negative datapoints in the dataset,
with the \emph{true positive rate} ($\mathrm{TPR}$) and \emph{false positive
  rate} ($\mathrm{FPR}$)  defined as
\begin{equation}
  \mathrm{TPR} = \frac{\mathrm{TP}}{\mathrm{TP} + \mathrm{FN}} = \frac{\mathrm{TP}}{\mathrm{P}}, \qquad
  \mathrm{FPR} = \frac{\mathrm{FP}}{\mathrm{FP} + \mathrm{TN}} = \frac{\mathrm{FP}}{\mathrm{N}}.
\end{equation}  

Let the fraction of positives in the dataset be denoted by $\pi =
\mathrm{P}/(\mathrm{P}+\mathrm{N})$, and define the ratio $r =
\mathrm{P}/\mathrm{N} = \pi/(1-\pi)$.
If we consider the table above normalized by the sample size
$n = \mathrm{P} + \mathrm{N}$, then we observe that the table's
entries are fully characterized by the three quantities
$\mathrm{TPR}$, $\mathrm{FPR}$
and $r$, as the sum of the normalized entries must be 1.
The values in the table are usually thought of as empirical counts
from a sample of size $n$. However, one can consider the normalized
table in the limit $n \rightarrow \infty$, which describes the
\emph{population} properties of the classifier at the threshold
chosen.

The ROC curve is a plot of $\mathrm{TPR}$ against $\mathrm{FPR}$.  As
is well known (see e.g., \citealt*{fawcett-06}), the population ROC is
invariant to $r$ ; this is immediate from the definitions of
$\mathrm{TPR}$ and $\mathrm{FPR}$, which are ratios within the
positives and negatives respectively. Empirical ROC curves for
will exhibit some variability as $r$ varies (and indeed across
different samples of the same size).


Precision is defined as
\begin{equation}
  \mathrm{Prec} = \frac{\mathrm{TP}}{\mathrm{TP} + \mathrm{FP}} =
  \frac{\mathrm{P} \cdot \mathrm{TPR}}{\mathrm{P} \cdot \mathrm{TPR} +
      \mathrm{N} \cdot \mathrm{FPR}}
      = \frac{\mathrm{TPR}}{\mathrm{TPR} + \frac{1}{r} \mathrm{FPR}}. \label{eq:prec}
\end{equation}
Thus the precision has an explicit dependence on $r$.
Note that the $\mathrm{Prec} \rightarrow 1$ as $\pi \rightarrow 1$, and
also that $\mathrm{Prec} \rightarrow 0$ as $\pi \rightarrow 0$ if
$\mathrm{FPR} > 0$.

\begin{figure}
\centering  
  \includegraphics[width=.7\textwidth]{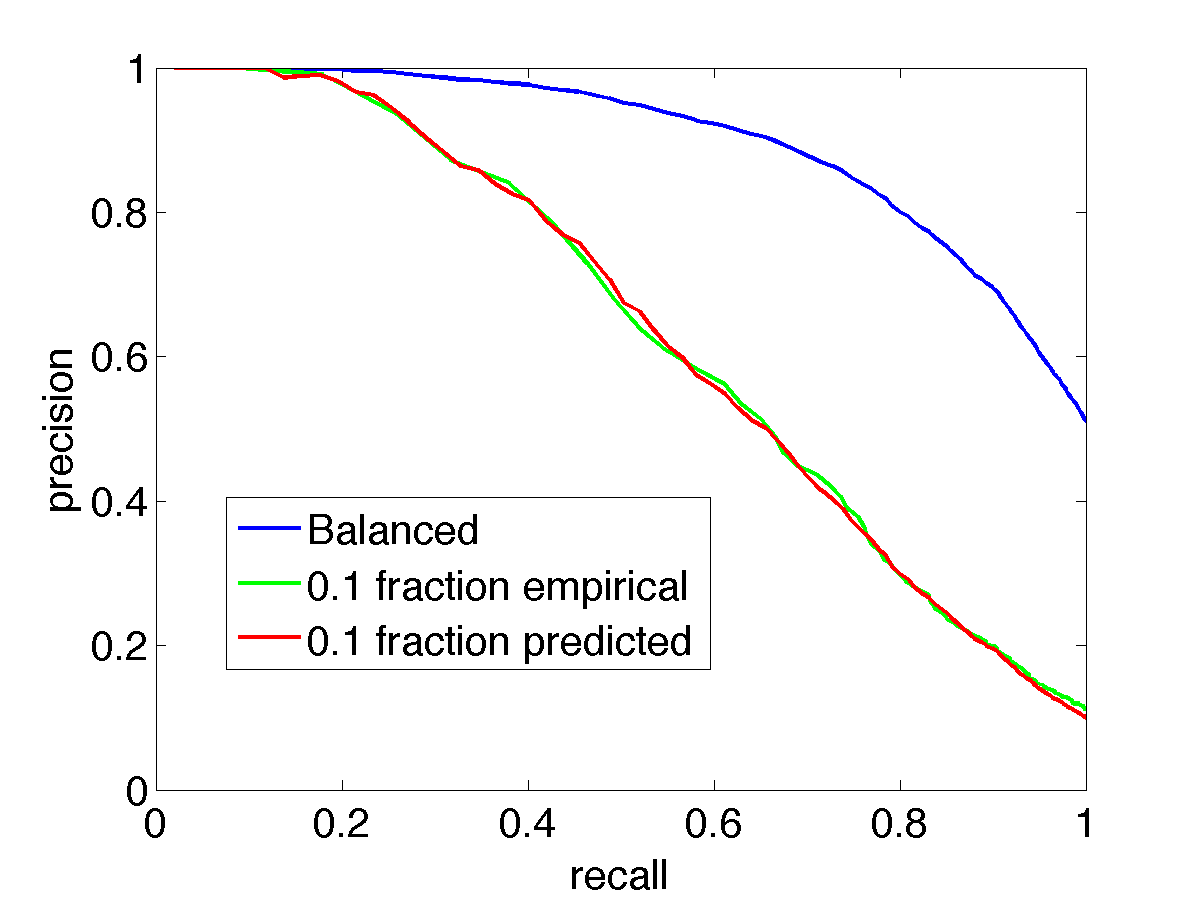}
\caption{Precision-recall curves for varying $r$.  \label{fig:PR}}
\end{figure}

The precision-recall curve plots the precision against recall
$\mathrm{Rec}$, which is another name for the true positive rate. As
recall is invariant to class imbalance, we can consider how the
precision varies with $r$ at fixed recall.  If we start with
balanced classes at $r=1$ and gradually decrease $r$\footnote{PR
  curves are typically used when $r$ is small, e.g.\ in information
  retrieval settings.}, we see that the corresponding precision will
decrease, because the denominator increases.

For population values of $\mathrm{TPR}$, $\mathrm{FPR}$ and $r$,
eq.\ \ref{eq:prec} allows us to transform the precision as a
function of $r$. For an empirical sample, it allows us to \emph{predict}
how the PR curve will change with $r$ using the empirical values
of $\mathrm{TPR}$ and $\mathrm{FPR}$. This is illustrated in Fig
\ref{fig:PR}. In this case a simple classification problem with 2d Gaussians
was set up, and a logistic regression classifier trained. For a
test set with $r = 1$ and $\mathrm{P} = \mathrm{N} = 5000$
the blue curve was obtained, and for $r = 0.1$ ($\mathrm{P} = 500$,
$\mathrm{N} = 5000$)
the green empirical curve. If at each value of recall the blue curve is scaled
as per eq.\ \ref{eq:prec}, the red curve is obtained. Note the good
agreement between the predicted and actual curves; the differences can be
explained by the fact that the empirical green curve uses a smaller
number of samples than the red curve (which reweights all of the
balanced samples).

The ability to predict how the PR curve varies with $r$ does not
seem to be well known. For example, \citet[sec~4.2]{fawcett-06}
discusses ``class skew'' and shows PR curves for $r=1$ and $r=0.1$,
but makes no comment on their relationship.
However, \citet{hoiem-chodpathumwam-dai-12} have pointed
out that when comparing PR curves for the detection of different
visual object classes, the average precision score is sensitive  to
the value of $r$ for each class. To enable a fairer comparison, they
suggested using ``normalized precision'', which uses a standard
value of $r$ across
classes\footnote{\citet{hoiem-chodpathumwam-dai-12}
  considered the PASCAL Visual Object Classes
  (VOC) dataset across 20 object classes, and chose their standard $r$
  based on the average proportion of positives across the classes.}.

Note that class imbalance $r_{\mathrm{train}}$ in the \emph{training}
data should not have an effect on the \emph{test} ROC and PR curves
of a probabilistic classifier\footnote{Or of one that provides a 
graded real-valued output, like a SVM.}. To see
this, consider the log odds ratio
\begin{equation}
\log \frac{p(C_+ | \bfx)}{p(C_- | \bfx)} = \log
\frac{p(\bfx|C_+)}{p(\bfx|C_-)}   + \log \rtrain ,  \label{eq:logodds}
\end{equation}
where $\rtrain = p(C_+)/p(C_-)$.  For a generative classifier the LHS
is obtained from the RHS and the effect of $\rtrain$ is immediate. For
a discriminative classifier eq.\ \ref{eq:logodds} can be used to
understand the effect of $\rtrain$ on the decision boundary. The test
ROC and PR curves only depend on the sequence of confusion matrices
obtained as the threshold on the classifier's log odds ratio is
changed---the effect of changes in $\rtrain$ is to shift the
threshold, but not to change the sequence obtained.

\vspace*{3mm}

The $F_{\beta}$ measure 
is commonly used as a figure-of-merit that
combines precision and recall. It is defined as a weighted harmonic average
\begin{equation}
  \frac{1}{F_{\beta}} = \frac{1}{1+ \beta^2} \frac{1}{\mathrm{Prec}}
    + \frac{\beta^2}{1+ \beta^2} \frac{1}{\mathrm{Rec}} .
\end{equation}
Substituting the expression for the precision from eq.\ \ref{eq:prec},
we obtain
\begin{equation}
  \frac{1}{F_{\beta}} = \frac{1}{1+ \beta^2} \frac{\mathrm{TPR} + \frac{1}{r} \mathrm{FPR}}{\mathrm{TPR}}
    + \frac{\beta^2}{1+ \beta^2} \frac{1}{\mathrm{TPR}},
\end{equation}
and hence
\begin{equation}
F_{\beta} = \frac{(1+\beta^2)\mathrm{TPR} }{\mathrm{TPR} + \frac{1}{r} \mathrm{FPR} +
  \beta^2},
\end{equation}
which demonstrates the explicit dependence of $F_{\beta}$ on $r$.

\vspace*{3mm}

The performance of a classifier is often summarized by the area under
the PR curve (AUPR), by analogy to the area under the ROC curve (AUROC).
However, \citet{flach-kull-15} argue that it is better to summarize
precision-recall performance based on the $F_1$ score.
This leads them to introduce the Precision
Gain $\mathrm{PrecG}$ and Recall Gain $\mathrm{RecG}$, defined as
\begin{equation}
  \mathrm{PrecG} =  \frac{\mathrm{Prec} - \pi }{(1-\pi) \mathrm{Prec}}, \qquad
  \mathrm{RecG} = \frac{\mathrm{Rec} - \pi }{(1-\pi) \mathrm{Rec}}.
\end{equation}
Their Precision-Recall-Gain curve plots Precision Gain on the y-axis
against Recall Gain on the x-axis in the unit square (i.e., negative
gains are ignored). It is interesting to express $\mathrm{PrecG}$ and
$\mathrm{RecG}$ in terms of  $\mathrm{TPR}$, $\mathrm{FPR}$ and
$r$. Using $1/(1 - \pi) = 1+r$ we obtain
\begin{align}
  \mathrm{PrecG} &= \frac{1}{1-\pi} - \frac{r}{\mathrm{Prec}} = 1 +r -
  r \left(1 + \frac{1}{r} \frac{\mathrm{FPR}}{\mathrm{TPR}} \right) = 1 -
  \frac{\mathrm{FPR}}{\mathrm{TPR}}, \\
  \mathrm{RecG} &= \frac{1}{1-\pi} - \frac{r}{\mathrm{Rec}} = 1 + r
  \left( 1  - \frac{1}{\mathrm{TPR}} \right) .
\end{align}  
Notice how $\mathrm{PrecG}$ is in fact independent of $r$, while
$\mathrm{RecG}$ has an affine rescaling due to $r$. Interestingly, both 
$\mathrm{PrecG}$ and $\mathrm{RecG}$ each only depend on two out
of the three quantities $\mathrm{TPR}$, $\mathrm{FPR}$ and $r$.

The key point of the above analyses is to highlight the explicit effect of the
class imbalance as expressed by $r$ on the precision, $F_{\beta}$ and the
precision/recall gains, and to show how these quantities can be
adjusted for different $r$ if necessary.
Like \citet{hoiem-chodpathumwam-dai-12}, \citet{siblini-etal-20}
make use of a fixed class
ratio $r_0$, and use it to define AUPR, F-score and AUPR Gain scores
that thus do not depend on $r$.

\subsection*{Acknowledgements}
I thank Nick Radcliffe for a question that started this work off,
Tom Dietterich for pointing out the work of \cite{flach-kull-15},
Peter Flach for pointing out a typo in eq.\ 6 in an earlier version,
Wissam Siblini for alerting me to \citet{siblini-etal-20}, 
the anonymous referees for comments that helped improve the paper, 
and Sim\~{a}o Eduardo, Alfredo Naz\'{a}bal and Charles Sutton for
helpful discussions.

\bibliographystyle{apalike}

\end{document}